\documentclass[]{elsarticle}
\usepackage[colorlinks=false,linkcolor=black, citecolor=black, urlcolor=black]{hyperref}

\usepackage{lineno,hyperref}
\modulolinenumbers[5]

\usepackage{tabularx} 
\usepackage{booktabs} 
\usepackage{amsmath}
\usepackage{multirow}
\usepackage{listings}
\usepackage{color}
\usepackage{subfig}
\usepackage{algorithm}
\usepackage{algpseudocode}
\definecolor{lightgray}{gray}{0.9}
\newcommand{\inlinecode}[1]{\colorbox{lightgray}{\lstinline|#1|}}

\newcolumntype{C}[1]{>{\centering\arraybackslash}m{#1}}

\journal{Arxiv}
\date{}

\begin{document}

\begin{frontmatter}

\title{Deep Transfer Learning for Source Code Modeling}






\author[add1]{Yasir Hussain\corref{cor1}}
\ead{yaxirhuxxain@nuaa.edu.cn}
\author[add1,add2,add3]{Zhiqiu Huang\corref{cor1}}
\ead{zqhuang@nuaa.edu.cn}
\author[add1]{Yu Zhou}
\ead{zhouyu@nuaa.edu.cn}
\author[add1]{Senzhang Wang}
\ead{szwang@nuaa.edu.cn}

\address[add1]{College of Computer Science and Technology, Nanjing University of Aeronautics and Astronautics (NUAA), Nanjing 211106, China}
\address[add2]{Key Laboratory of Safety-Critical Software, NUAA, Ministry of Industry and Information Technology, Nanjing 211106, China}
\address[add3]{Collaborative Innovation Center of Novel Software Technology and Industrialization, Nanjing 210093, China}

\cortext[cor1]{Corresponding author}

\begin{abstract}
	In recent years, deep learning models have shown great potential in source code modeling and analysis. Generally, deep learning-based approaches are problem-specific and data-hungry. A challenging issue of these approaches is that they require training from starch for a different related problem. In this work, we propose a transfer learning-based approach that significantly improves the performance of deep learning-based source code models. In contrast to traditional learning paradigms, transfer learning can transfer the knowledge learned in solving one problem into another related problem. First, we present two recurrent neural network-based models RNN and GRU for the purpose of transfer learning in the domain of source code modeling. Next, via transfer learning, these pre-trained (RNN and GRU) models are used as feature extractors. Then, these extracted features are combined into \textit{attention} learner for different downstream tasks. The \textit{attention} learner leverages from the learned knowledge of pre-trained models and fine-tunes them for a specific downstream task. We evaluate the performance of the proposed approach with extensive experiments with the source code suggestion task. The results indicate that the proposed approach outperforms the state-of-the-art models in terms of accuracy, precision, recall, and F-measure without training the models from scratch.
\end{abstract}


\end{frontmatter}

\section{Introduction}
Source code suggestion and syntax error fixing are vital features of a modern integrated development environment (IDE). These features help software developers to build and debug software rapidly. Recently, deep learning-based language models have shown great potential in various source code modeling tasks \cite{allamanis2016convolutional,alon2019code2vec,santos2018syntax,gupta2018deep,iyer2016summarizing,fowkes2017autofolding,raychev2014code,sethi2018dlpaper2code,white2015toward,hussain2019deepvs}. Several studies have explored deep learning for source code suggestion \cite{raychev2014code,white2015toward} in which they suggest the next possible source code token. They take a fixed size context prior to the prediction position as features and help the software developers by suggesting the next possible code token. Further, deep learning has recently been explored for syntax error detection and correction \cite{santos2018syntax,gupta2018deep} problem. They consider the source code syntax as features and use them for the correction of the syntax errors found in a source code file. Moreover, deep learning has shown its effectiveness in the source code summarization \cite{allamanis2016convolutional,iyer2016summarizing,fowkes2017autofolding} , which summarizes the working of source code. 

A challenging issue of these approaches is that they are problem-specific which requires training from starch for a different related problem. Further, deep learning-based approaches are data-hungry which means they require training on large data set to produce satisfactory results. Furthermore, deep learning models requires days to train while training on a large dataset. To overcome these issues, we exploit the concept of transfer learning in this work. In transfer learning, the learned knowledge from a pre-trained model is extracted and then be used for a similar downstream task \cite{salem2019utilizing}.

This work proposes a transfer learning-based approach that significantly improves the performance of deep learning-based source code models. First, we exploit the concept of transfer learning for deep learning-based source code language models. The key idea is to use a pre-trained source code language model to transfer the learned knowledge from it to a different related problem. We train two different variants of recurrent neural network-based models RNN and GRU for the purpose of transfer learning. Then, we combine the learned knowledge of pre-trained (RNN and GRU) models into \textit{attention} learner for a downstream task. The \textit{attention} learner leverage from the learned knowledge of pre-trained models and fine-tunes it for a specific downstream task. Via transfer learning, pre-trained models are used to extract generalized features and then fine-tune them for a target task without requiring the model training from scratch. We evaluate the proposed approach with the downstream task of source code suggestion.

This work makes the following unique contributions:

\begin{itemize}
	
	\item We exploit the concept of transfer learning in the domain of source code. We propose transfer learning-based \textit{attention} learner approach for the downstream task of source code suggestion. 
	
	\item We present two recurrent neural network-based (RNN and GRU) pre-trained models for the purpose of transfer learning in the domain of source code.
	
	\item An extensive evaluation of the proposed approach on the real-world data set shows significant improvement in the state-of-the-art language models.
	
\end{itemize}

\section{Related Work}
In this section, we present background study on deep learning, transfer learning and source code language models.

\subsection{Source Code Modeling}
Hindle et al. \cite{hindle2012naturalness} have shown how natural language processing techniques can help in source code modeling. They provide a \textit{n-gram} based model which helps predict the next code token in \textit{Eclipse IDE}.  Tu et al. \cite{tu2014localness}, proposed a cache-based language model that consists of an \textit{n-gram} and a \textit{cache}. Hellendoorn et al. \cite{hellendoorn2017deep} further improved the cache-based model by introducing nested locality. Another approach for source code modeling is to use probabilistic context-free grammars(PCFGs) \cite{bielik2016phog}. Allamanis et al. \cite{allamanis2014mining} used a PCFG based model to mine idioms from source code. Maddison et al. \cite{maddison2014structured} used a structured generative model for source code. They evaluated their approach with \textit{n-gram} and \textit{PCFG} based language models and showed how they can help in source code generation tasks. Raychev et al.\cite{raychev2016learning} applied decision trees for predicting API elements. Chan et al. \cite{chan2012searching} used a graph-based search approach to search and recommend API usages.

Recently there has been an increase in API usage \cite{wang2013mining,keivanloo2014spotting,d2016collective} mining and suggestion. Thung et al. \cite{thung2013automatic} introduced a recommendation system for API methods recommendation by using feature requests. Pham et al. \cite{pham2016learning} proposed a methodology to learn API usages from byte code. Hussain et al. \cite{hussain2019codegru} proposed GRU based model for source code suggestion and completion task (completion of a whole line of code). A neural probabilistic language model introduced in \cite{allamanis2015suggesting} that can suggest names for the methods and classes. Franks et al. \cite{franks2015cacheca} created a tool for Eclipse named \textit{CACHECA} for source code suggestion using a \textit{n-gram} model. Nguyen et al. \cite{nguyen2012graph} introduced an \textit{Eclipse plugin} which provide code completions by mining the API usage patterns. Chen et al. \cite{chen2016similartech} created a web-based tool to find analogical libraries for different languages.

A similar work conducted by Rabinovich et al. \cite{rabinovich2017abstract}, which introduced an abstract syntax networks modeling framework for tasks like code generation and semantic parsing. Sethi et al. \cite{sethi2018dlpaper2code} introduced a model which automatically generate source code from deep Learning-based research papers. \cite{allamanis2015bimodal}, Allamanis et al. proposed a bimodal to help suggest source code snippets with a natural language query. Recently deep learning-based approaches have widely been applied for source code modeling. Such as code summarization \cite{iyer2016summarizing,allamanis2016convolutional,guerrouj2015leveraging,jiang2017prst}, code mining \cite{xie2006mapo}, clone detection \cite{kumar2015code,zhang2015event},  API learning \cite{gu2016deep,wang2014mmrate}, code generation \cite{zhou2019augmenting} etc.

We observe \cite{raychev2014code,white2015toward} approaches are related to our downstream task of source code suggestion. Raychev et al. \cite{raychev2014code} used RNN for the purpose of code completion specifically focusing on suggesting source code method calls. Similarly, White et al. \cite{white2015toward} applied RNN based deep neural network for source code completion task. Generally, these approaches \cite{raychev2014code,white2015toward} are problem-specific which requires training from scratch for a different related problem. In this work, we exploit the concept of transfer learning to extract the learned knowledge from pre-trained models and then fine-tunes it for a related problem, which shows a significant performance boost, without requiring the model’s training from scratch.

\subsection{Transfer Learning}
Transfer learning as the name suggests intending to transfer knowledge (features) learned in solving one problem into another related problem. Hu et al. \cite{hu2015deep} have proposed a transfer metric learning approach for visual recognition in cross-domain datasets. Duan et al. \cite{duan2009domain} have proposed a kernel learning approach for the detection of cross-domain keyframe feature changes. Pan et al. \cite{pan2008transfer} have proposed a dimensionality reduction method which uses the transfer learning approach by minimizing the distance between distributions between target and source domains. Khan et al. \cite{khan2019novel} have proposed a deep transfer learning approach for the detection of breast cancer by using pre-trained GoogLeNet, VGGNet, and ResNet. Huang et al. \cite{huang2017transfer} have proposed a transfer learning-based approach for Synthetic Aperture Radar (SAR) classification with limited labeled data. Kraus et al. \cite{kraus2017decision} proposed a decision support system by using deep neural networks and transfer learning for financial disclosures. Further, transfer learning has been extensively studied for various tasks in the field of image and text classification \cite{shin2016deep,zhang2016deep,yuan2017hyperspectral,kraus2017decision}. In this work, we exploit the transfer learning approach for the purpose of source code modeling. Instead of using a single model for transferring knowledge, in this work, we use a novel approach that transfers knowledge from two different recurrent neural network-based pre-trained models and then fine-tunes it by using attention learner for a specific source code modeling tasks.

\section{Preliminary } \label{preliminary }
In this section, we will discuss the preliminaries and technical overview of this work.

\subsection{Recurrent Neural Network}

The recurrent neural network has recently shown great potential in a wide range of applications including image recognition, text classification, and source code modeling. However, a major drawback of vanilla RNN is the vanishing gradient which can be overcome by using the gated recurrent unit (GRU) \cite{young2018recent}. The GRU exposes the full hidden content on each timestep, thus evading the disappearing gradient issue. It can be expressed as
\begin{equation}
h_i = (1-z_i)h_{i-1}+z_i\hbar_i
\end{equation}

\begin{equation}
z_i = \phi(W_z\tau_i+U_zh_{i-1})
\end{equation}

\begin{equation}
\hbar_i = tanh(W\tau_i+r_i \otimes Uh_{i-1})
\end{equation}

\begin{equation}
r_i = \phi(W_r\tau_i+U_rh_{i-1})
\end{equation}

\subsection{Attention Learner}
Recently, \textit{attention}-based approaches have shown great potential in various fields such as speech recognition \cite{chorowski2015attention}, machine translation \cite{luong2015effective,bahdanau2016end}, and more \cite{vaswani2017attention,alon2019code2vec}. The attention (\cite{bahdanau2016end}) model calculates a context vector as the weighted mean of the state sequence. it can be expressed as

\begin{equation}
a_i^t = \frac{exp(c_i^t)}{\sum_{j=1}^{T}exp(c_j^t)}
\end{equation}
\begin{equation}
c_i^t = a(s^{t-1},h_i)
\end{equation}

\section{Proposed approach} \label{Methodology}
\begin{figure*}[htbp]
	\centering
	\includegraphics[width=\linewidth]{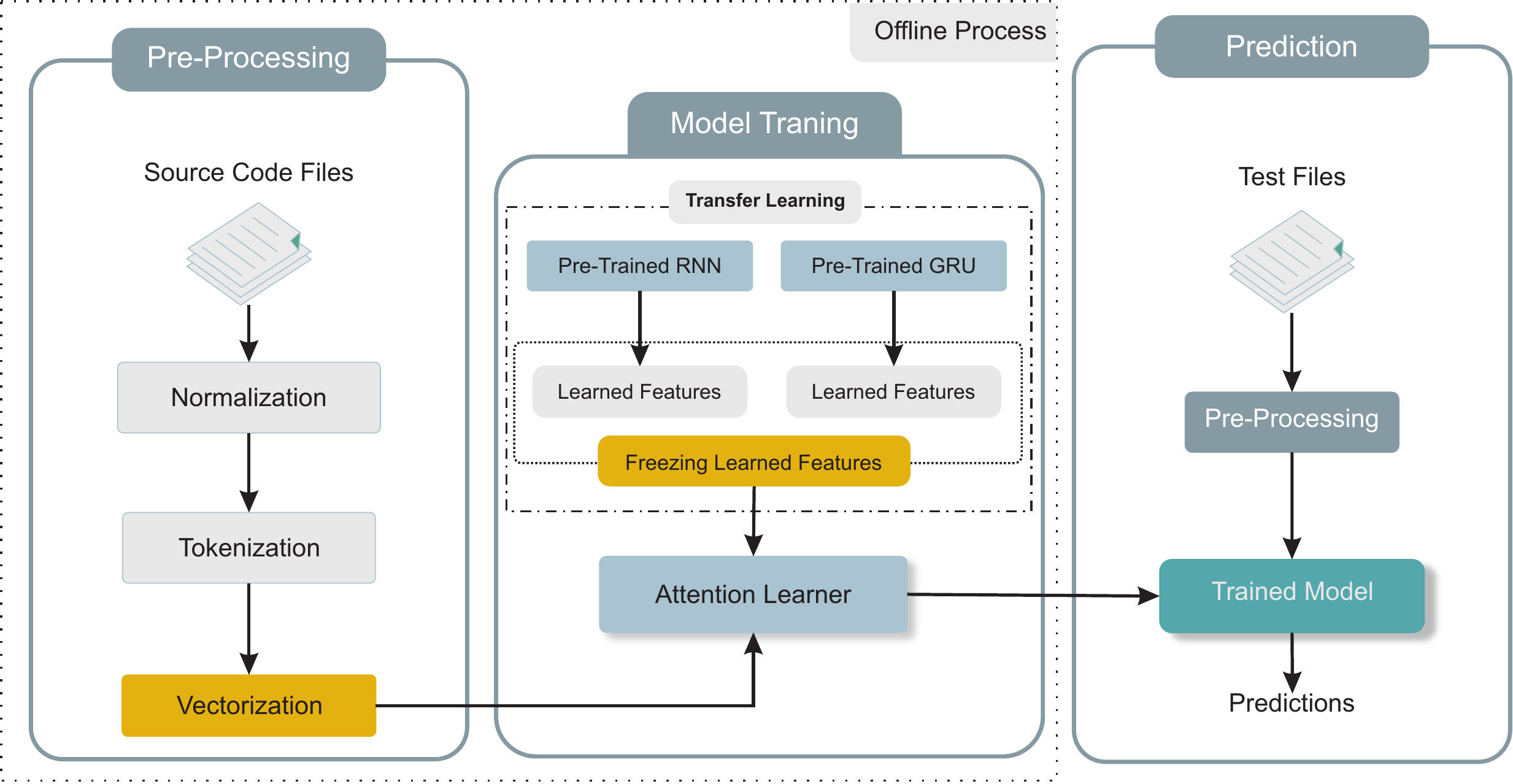}
	\caption{Overall framework of the proposed approach}
	\label{fig:framework}
\end{figure*}

This section discusses the proposed approach in detail. The \hyperref[fig:framework]{Fig.} \ref{fig:framework} shows the overall workflow of the proposed approach. This section is subdivided into two major parts. In the first part, we discuss the preparation of pre-trained models for the purpose of transfer learning. The second part discusses the key steps involved to prepare the source code for the downstream task of source code suggestion. Further, we discuss the attention learner, which leverages the learned knowledge from pre-trained models and fine-tunes it for the source code suggestion task. The details about each step are given in the following subsections.

\subsection{Transfer Learning}\label{TransferLearning}
\begin{figure*}[t!]
	\centering
	\subfloat[]{\includegraphics[width=0.37\linewidth]{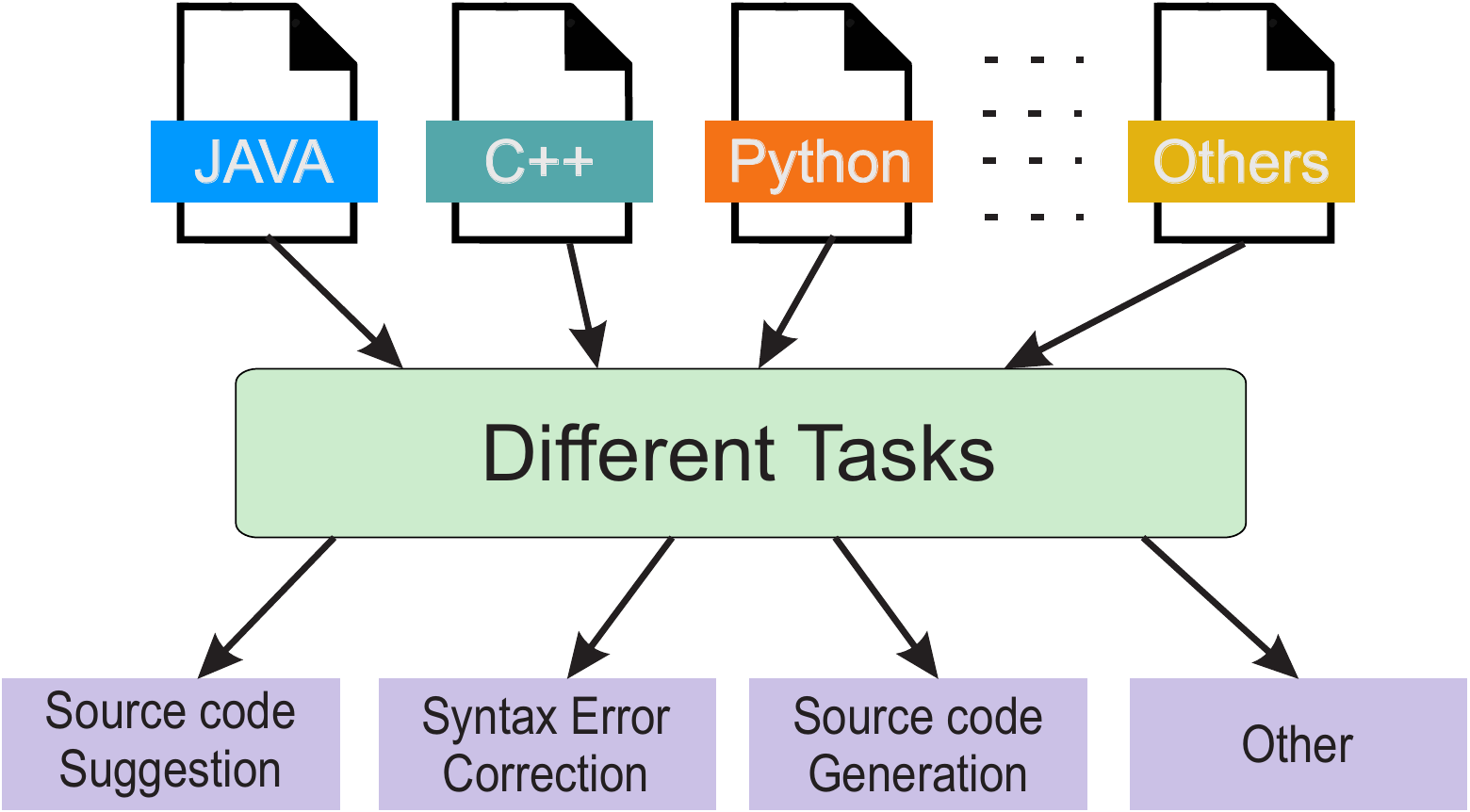}}
	\subfloat[]{\includegraphics[width=0.6\linewidth]{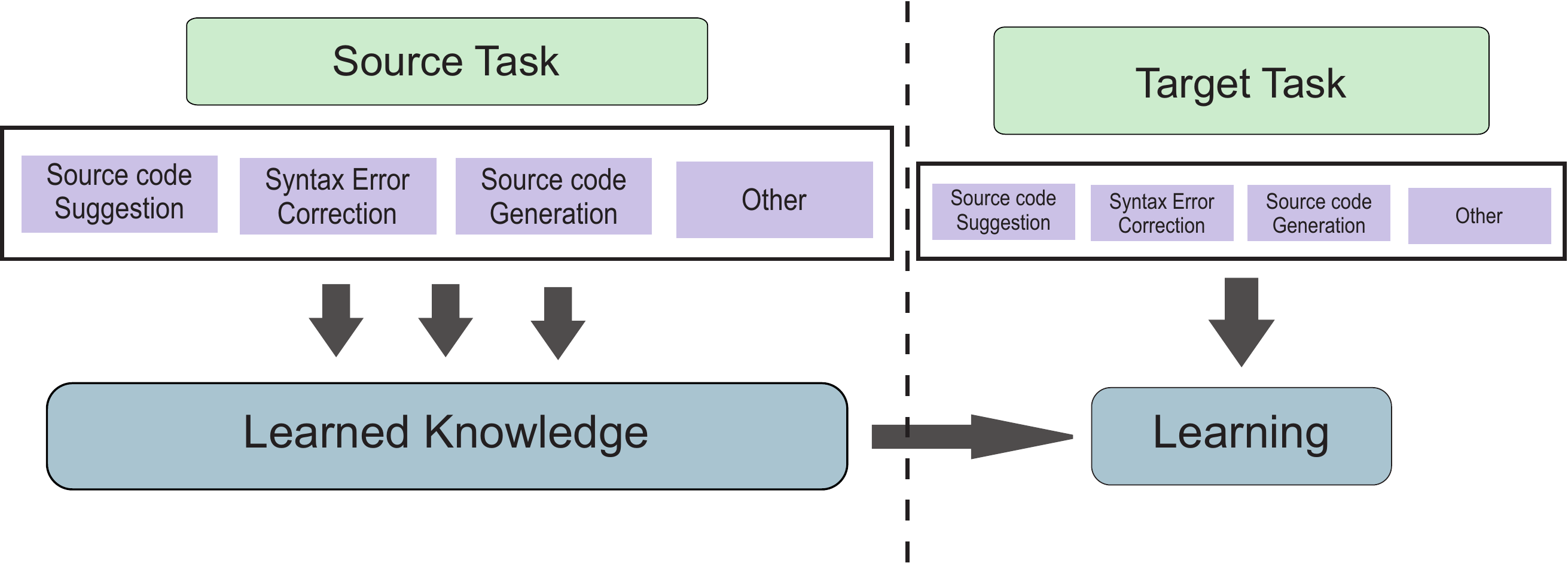}}
	\caption{Difference between traditional learning and transfer learning based approaches. (a) Traditional learning approach; (b) Transfer learning based approach.}
	\label{fig:learningComparision}
\end{figure*}

\hyperref[fig:learningComparision]{Fig.} \ref{fig:learningComparision} shows the difference between traditional learning and transfer learning-based approaches for source code modeling. For the purpose of transfer learning, we first need a pre-trained model. There are several CNN \cite{szegedy2015going,he2016deep} and NLP \cite{devlin2018bert,dai2019transformer,radford2019language} based models for image and text classification respectively. The source code strictly follows the rules defined by their grammar\footnote{\url{https://docs.oracle.com/javase/specs/jls/se7/html/jls-18.html}}, thus these models are not suitable for our purpose. In this work, we first train two variants of recurrent neural networks-based models RNN and GRU for the purpose of transfer learning in the field of source code. We choose RNN and GRU because of their recent success in the modeling of source code. To train the models for transfer learning, we gather the data set used in previous studies \cite{hindle2012naturalness,nguyen2018deep}. \hyperref[Table:TopDataStatistics]{Table} \ref{Table:TopDataStatistics} shows the details of the data set used to build pre-trained models. By combining all collected projects, we end up with 13 million code tokens with a large vocabulary of size 177,342. 

\begin{table}[htpb]
	\small
	\caption{Data set used to pre-train models for transfer learning. The table shows name of the project, version of the project, line of code (LOC), total code tokens and unique code tokens found in each project.}
	\label{Table:TopDataStatistics}
	\begin{center}
		\begin{tabular}{ccccc}
			\toprule
			{Projects} & {Version} & {LOC} &{Total} & {Vocab Size ($V$)}\\
			\midrule
			ant & 1.10.5 & 149,960 & 920,978 &  17,132\\
			cassandra & 3.11.3 & 318,704 & 2734218 &  33,424\\
			db40 & 7.2  & 241,766 & 1,435,382 &  20,286\\
			jgit & 5.1.3  & 199,505 & 1,538,905 &  20,970\\
			poi & 4.0.0  & 387,203 & 2,876,253 &  47,756\\
			maven & 3.6.0  & 69,840 & 494,379 &  8,066\\
			batik & 1.10.0  & 195,652 & 1,246,157 &  21,964\\
			jts & 1.16.0  & 91,387 & 611,392 &  11,903\\
			itext & 5.5.13 & 161,185 & 1,164,362 &  19,113\\
			antlr & 4.7.1 & 56,085 & 407,248 &  6,813\\
			\midrule
			Total & & {1,871,287} & {13,429,274} & {177,342} \\
			\bottomrule
		\end{tabular}
	\end{center}
\end{table}

\subsubsection{Pre-Training Models for Transfer Learning} \label{MethodologyTraining}
All models are trained on Intel(R) Xeon(R) Silver 4110 CPU \@ 2.10GHz x 32 cores and 128GB of ram running Ubuntu 18.04.2 LTS operating system, equipped with the latest NVIDIA GeForce RTX 2080. The \hyperref[Table:DeepModelsArch]{Table.} \ref{Table:DeepModelsArch} shows the architecture of trained models used for transfer learning. We follow the same approach used in previous works \cite{white2015toward,raychev2014code} to pre-process the data set. To build a global vocabulary system, we remove all code tokens appearing less than three times in the collected data set which ends up with the vocabulary size of 88,022 unique code tokens. We map the vocabulary ($V$) to a continuous feature vector of dense size \textit{300}. We use \textit{300} hidden units with context size ($\tau$) of \textit{20}. For each model training we employ \textit{Adam} optimizer with the default learn rate of \textit{0.001}. To control overfitting, we use \textit{Dropout}. Each model is trained until it converges by employing \textit{early stop} with the patience of three consecutive hits on the validation loss. One important thing to point out here is that the training process of these models is one time and do not need retraining. The trained models are publicly available\footnote{Trained Models: \url{https://github.com/yaxirhuxxain/TransferLearning}} for the purpose of transfer learning.

\begin{table}[!t]
	\small
	\caption{Deep learning models architecture summary used to pre-train source code models for transfer learning purpose.}
	\label{Table:DeepModelsArch}
	\begin{center}
		\begin{tabular}{*{5}{r}}
			\toprule
			& Type & Size & Activations  \\
			\midrule
			Input & Code embedding & 300 &  &  \\
			Estimator & RNN,GRU & 300 & tanh & \\
			Over Fitting & Dropout & &  & \\
			Output & Dense & $V$  & softmax  & \\
			Loss & Categorical cross entropy& &  & \\
			Optimizer & Adam &  &  & \\
			\bottomrule
		\end{tabular}
	\end{center}
\end{table}

\subsection{Learning to Transfer Knowledge}
\begin{figure*}[htbp]
	\centering
	\includegraphics[width=\linewidth]{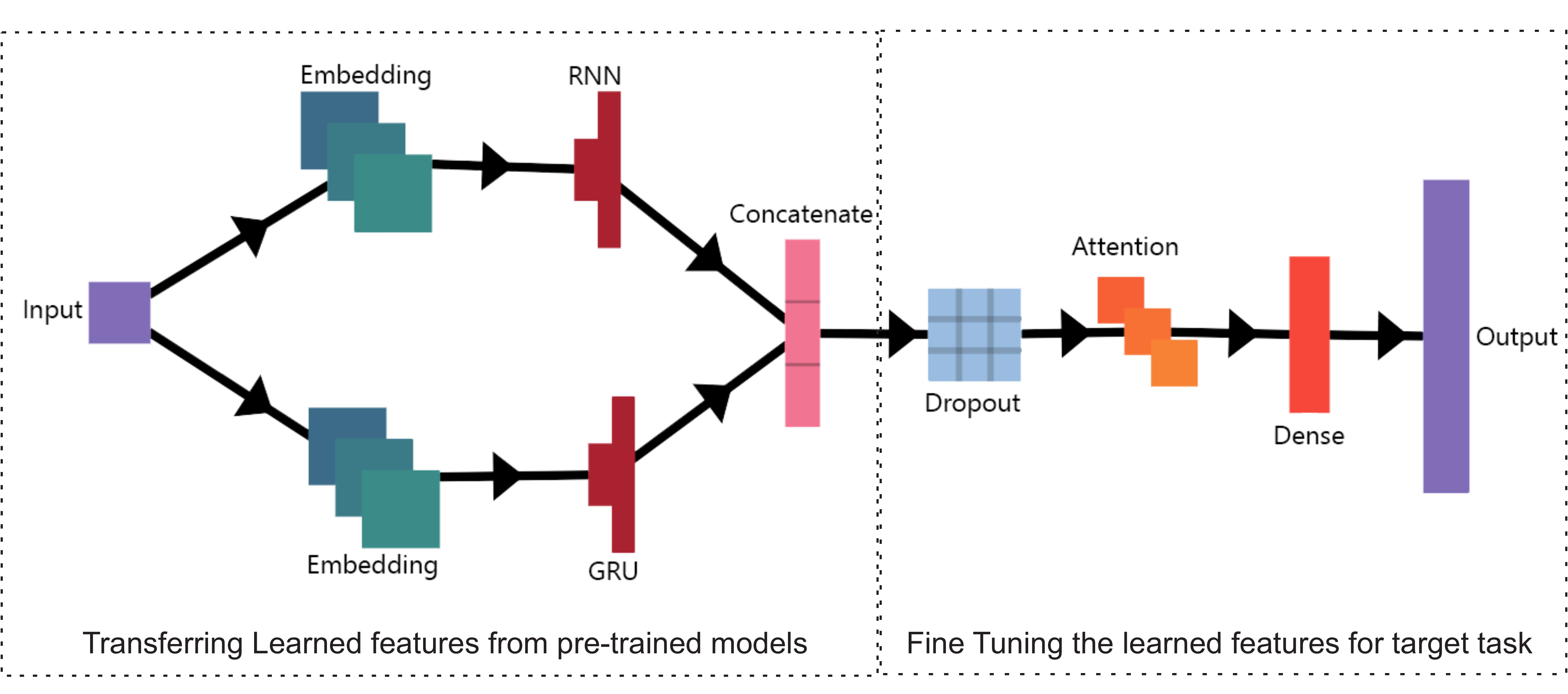}
	\caption{Proposed transfer learning based attention model architecture.}
	\label{fig:TransferLearnArch}
\end{figure*}

For transfer learning, we prepared the pre-trained models as described earlier in this section. Then, we use these pre-trained models to transfer the learned knowledge for the downstream task of source code suggestion. A key insight is to freeze the learned knowledge in pre-trained models to keep the learned knowledge unchanged and then fine-tune the extracted knowledge. In the proposed approach, we use the \textit{attention} learner to fine-tune the model for the source code suggestion task. The \textit{attention} learner pays attention to the task-specific features to achieve optimal performance. The \hyperref[fig:TransferLearnArch]{Fig.} \ref{fig:TransferLearnArch} shows the architecture design of our proposed transfer learning-based \textit{attention} model. We show the effectiveness of the proposed approach with the downstream task of source code suggestion. A source code suggestion engine recommends the next possible source code token given a context.

\subsubsection{PreProcessing} \label{PreProcessing}
In this section, we briefly introduce each of the key preprocessing steps that we apply for the downstream task of source code suggestion. We perform normalization, tokenization and feature extraction. For the illustrating example, \hyperref[Table:Preprocessing]{Table} \ref{Table:Preprocessing} shows the effect of each preprocessing step. We discuss each step in detail in the following subsections.

\begin{table*}[htbp]
	\small
	\caption{An example of preprocessing steps.}
	\label{Table:Preprocessing}
	\begin{center}
		\begin{tabular}{ C{3.5cm} | C{8cm} }
			\toprule
			Original Source code & \includegraphics[width=0.7\linewidth]{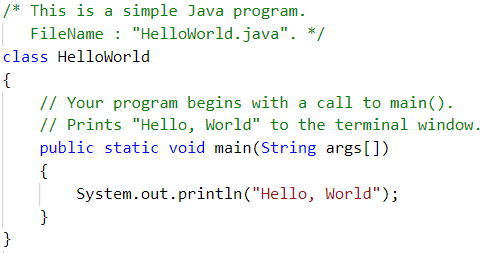}\\
			\hline \\
			
			Normalized Source Code & \includegraphics[width=0.6\linewidth]{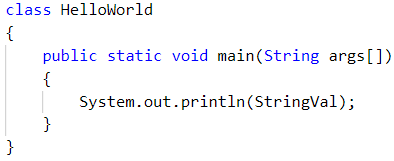}\\
			\hline \\
			
			Tokenized Source Code & \includegraphics[width=\linewidth]{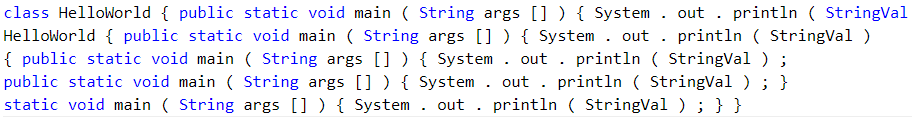}\\
			\hline \\
			
			Vectorized Source Code & \includegraphics[width=0.7\linewidth]{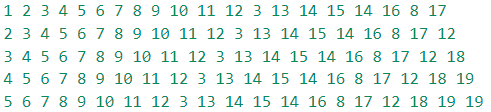}\\

			\bottomrule
		\end{tabular}
	\end{center}
\end{table*}

\subsubsection*{Normalization}
One of the vital preprocessing steps is to normalize the data set. Usually, a data set contains some values which are unnecessary for a particular task, these type of values will intensely upset the outcome of the analysis. For this purpose, we normalize the source code files by removing all blank lines, inline and block-level comments. We replace all constant numerical values to their generic types (e.g. 1 = IntVal, 1.2 = FloatVal) and replace constant strings with a generic \textit{StringVal} token.

\subsubsection*{Tokenization}
After normalizing source code files, we tokenize the source code files. Tokenization is the process of extracting terms/words from the data set. For this purpose, each source code file is parsed into a sequence of space-separated code tokens. Each sequence is then parted into multiple subsequences of fixed size context \textit{20} \cite{white2015toward}.

\subsubsection*{Vectorization}
To convert the source code sequences into a form that is suitable for training deep learning models we perform a series of transformations. First, we replace common tokens occurring only once in the corpus with a special token \textit{unk} to build a global vocabulary system. Next, we build the vocabulary where each unique source code token corresponds to an entry in the vocabulary. Then each source code token is assigned a unique positive integer corresponding to its vocabulary index to convert the sequences (feature vectors) into a form that is suitable for training a deep learning model.

\begin{table*}[htbp]
	\small
	\caption{Proposed transfer learning based attention model architecture summary.}
	\label{Table:DeepAttantionModelsArch}
	\begin{center}
		\begin{tabular}{c|c|c|c|c}
			\toprule
			& Layers & Type & Size & Activations  \\
			\midrule
			\multirow{3}{*}{Frozen} & Input & Code embedding & 300 &   \\
			& Estimator & RNN,GRU & 300 & tanh \\
			& Combining & Concatenate &  &  \\
			\midrule
			\multirow{4}{*}{Fine Tuned} & Over Fitting & Dropout & &  \\
			& Attention & Attention Learner &   &   \\
			& Output & Dense & $V$  & softmax  \\
			& Loss & Categorical cross entropy& &  \\
			& Optimizer & Adam &  &  \\
			\bottomrule
		\end{tabular}
	\end{center}
\end{table*}

\section{Evaluation} \label{Evaluation}
In this section, we evaluate the effectiveness of our proposed approach by investigating the following research questions:

\begin{itemize}
	\item RQ1: Does the proposed approach outperform the state-of-the-art approaches? if yes, to what extent?
	\item RQ2: How well the proposed approach performs in terms of source code suggestion task as compared to other baseline approaches?
	\item RQ3: Does normalization help to improve the performance of the proposed approach? If yes, to what extent?
\end{itemize}

To answer the research question (RQ1), we compare the performance of the proposed approach with the state-of-the-art approaches. To answer the research question (RQ2), we evaluate and compare the proposed approach for source code suggestion tasks with other baseline approaches. To answer the research question (RQ3), We conduct a comparative analysis to show the impact of normalization on model performance. 

\subsubsection{Data set} \label{Dataset}
To empirically evaluate our work, we collected java projects from \textit{GitHub} a well-known open-source software repositories provider. We gather the top five java projects sorted by the number of stars from \textit{GitHub} at the time of this study. We download the latest snapshot of the project usually named as the \textit{master branch}. Here, we choose the projects which are not used while training the pre-trained models discussed in \hyperref[Methodology]{Section} \ref{Methodology}. The \hyperref[Table:DataSet]{Table} \ref{Table:DataSet} shows the version of each project, the total number of code lines, total code tokens and unique code tokens found in each project. To empirically evaluate our work, we repeat our experiment on each project separately. We randomly partition the projects into ten equal lines of code folds from which one fold is used for testing, one fold is used for model parameter optimization (validation) and rest of the folds are used for model training.

\begin{table*}[h]
	\small
	\caption{List of projects used for the evaluation of this work.}
	\label{Table:DataSet}
	\centering
	\begin{center}
		\begin{tabular}{*{6}{r}}
			\toprule
			&  &  &  \multicolumn{2}{c}{Code Tokens}\\
			{Projects} & {Version} &{LOC} &{Total} & {Vocab Size ($V$)}\\
			\midrule
			elastic-search & v7.0.0 & 210,357 & 1,765,479 &  24,691\\
			java-design-patterns  & v1.20.0  & 30,784 & 200,344 &  5,649\\
			RxJava & v2.2.8  & 257,704 & 1,908,258 &  12,230\\
			interviews & v1.0  & 13,750 & 80,074 &  1,157\\
			spring-boot & v2.2.0.M2 & 224,465 & 1,813,891 &  34,609\\
			\bottomrule
		\end{tabular}
	\end{center}
\end{table*}

\subsection{Training and Prediction}

We train several baseline models for the evaluation of this work. The proposed approach is evaluated in the following manner; 

\begin{itemize}
	\item We train RNN \cite{raychev2014code} based model as baseline similar to White et al. \cite{white2015toward}.
	\item We train GRU based deep neural model as baseline similar to Cho et al.\cite{cho2014learning}
	\item We train transfer learning-based \textit{attention} model by following the proposed approach as discussed in \hyperref[Methodology]{Section} \ref{Methodology}.
\end{itemize}

We choose the approach proposed by White et al. \cite{white2015toward} for comparison because they have shown the effectiveness of their approach with the similar task of source code suggestion and as far as we know, considered as the state-of-the-art approach. We train the GRU \cite{cho2014learning} based model as the baseline because GRU based model is an advanced version of RNN which removes the vanishing gradient problem and performs better. The \hyperref[Table:DeepAttantionModelsArch]{Table.} \ref{Table:DeepAttantionModelsArch} shows the proposed transfer learning-based attention model architecture. First, we preprocess the data set as discussed earlier in \hyperref[PreProcessing]{Section} \ref{PreProcessing}. Then, we map the vocabulary to a continuous feature vector of dense size \textit{300}. We use \textit{300} hidden units with context size ($\tau$) of \textit{20}. For each model training we employ \textit{Adam} optimizer with the default learn rate of \textit{0.001}. To control overfitting, we use \textit{Dropout}. Each model is trained until it converges by employing \textit{early stop} with the patience of three consecutive hits on the validation loss.

For the prediction of next source code token \textit{y} in a source code file, the model takes the context information prior to the prediction position \textit{y}. Then, we use the trained models to predict the most likely next source code suggestions for the given context. If the predicted source code token is the same one as the original then we consider it a success.

\subsection{Metrics}
Usually deep learning approaches are evaluated by using different performance metrics. We choose similar evaluation approach as in previous studies. We choose top-k accuracy \cite{white2015toward,raychev2014code} and Mean Reciprocal Rank (MRR) \cite{nguyen2018deep,santos2018syntax} metrics for the evaluation of this work. Further, to evaluate the performance of the proposed approach we measure the precision, recall and F-measure scores which are widely used metrics \cite{alon2019code2vec}. Furthermore, to evaluate the significance of the proposed approach we perform ANOVA statistical testing. The computed metrics are formalized as,

\begin{equation}\label{eq:accuracy}
Accuracy = \frac{TP+TN}{TP+FN+FP+TN}
\end{equation}
\begin{equation}\label{eq:precision}
Precision = \frac{TP}{TP+FP}
\end{equation}
\begin{equation}\label{eq:recall}
Recall = \frac{TP}{TP+FN}
\end{equation}
\begin{equation}\label{eq:fmeaure}
\text{F-measure} = 2 \ast \frac{Precision\ast Recall}{Precision+Recall}
\end{equation}

Where true positive (TP) defines the total number of source code suggestions that are predicted correctly by the model. The true negative (TN) defines the total number of source code suggestions that are predicted incorrectly by the model. The false positive (FP) defines the total number of source code suggestions that are mistakenly predicted correctly by the model. Similarly, the false negative (FN) defines the total number of source code suggestions that are mistakenly predicted incorrectly by the model.

\section{Results} 
In this section, we will discuss and compare the results of our proposed approach with other baseline models.

\subsection{RQ1: Comparison against the baseline approaches}
\begin{figure}[h]
	\centering
	\includegraphics[width=\linewidth]{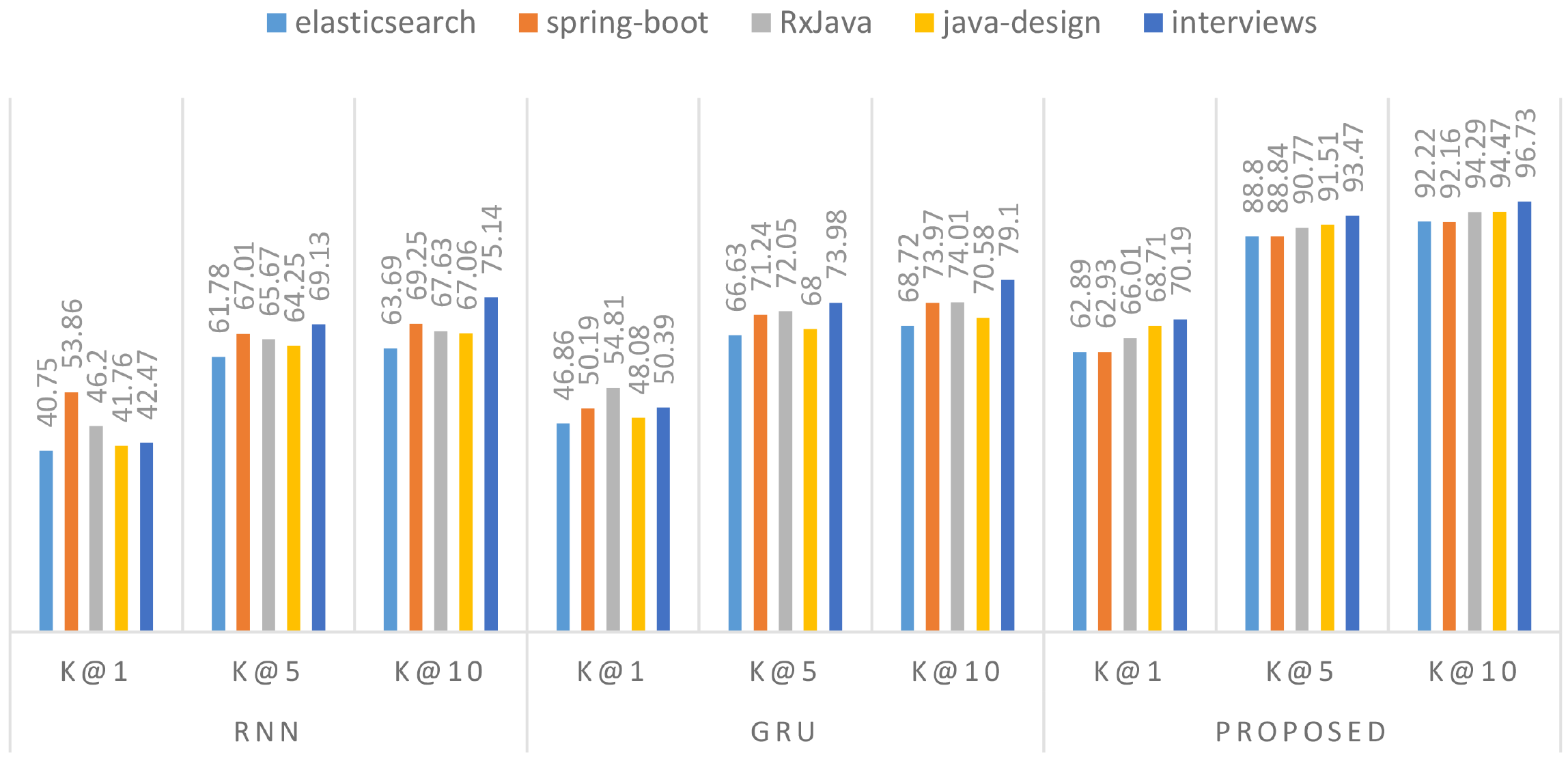}
	\caption{Top-k accuracy comparison.}
	\label{fig:top-k-accuracy}
\end{figure}

The top-k accuracy score of the proposed approach and the baseline approaches are presented in \hyperref[fig:top-k-accuracy]{Fig.} \ref{fig:top-k-accuracy}.  We make the following observations form \hyperref[fig:top-k-accuracy]{Fig.} \ref{fig:top-k-accuracy}

\begin{itemize}
	\item The average accuracy rate of RNN based model is \textit{45.01\%@k=1}, \textit{65.56\%@k=5} and \textit{68.55\%@k=10}, for the GRU based model is \textit{50.06\%@k=1}, \textit{64.38\%@k=5} and \textit{73.27\%@k=10}, while the proposed approach's average score is \textit{66.15\%@k=1}, \textit{90.68\%@k=5} and \textit{93.97\%@k=10} which is much higher as compared to the baseline approaches.
	\item On average the proposed approach improves the accuracy \textit{(k@1)}  by \textit{21.14\% } from RNN and \textit{16.09\%} from GRU based model.
	\item Results suggests that by employing the transfer learning-based \textit{attention} model it significantly improves the model performance.  
\end{itemize}

\begin{figure}[h]
	\centering
	\includegraphics[width=0.8\linewidth]{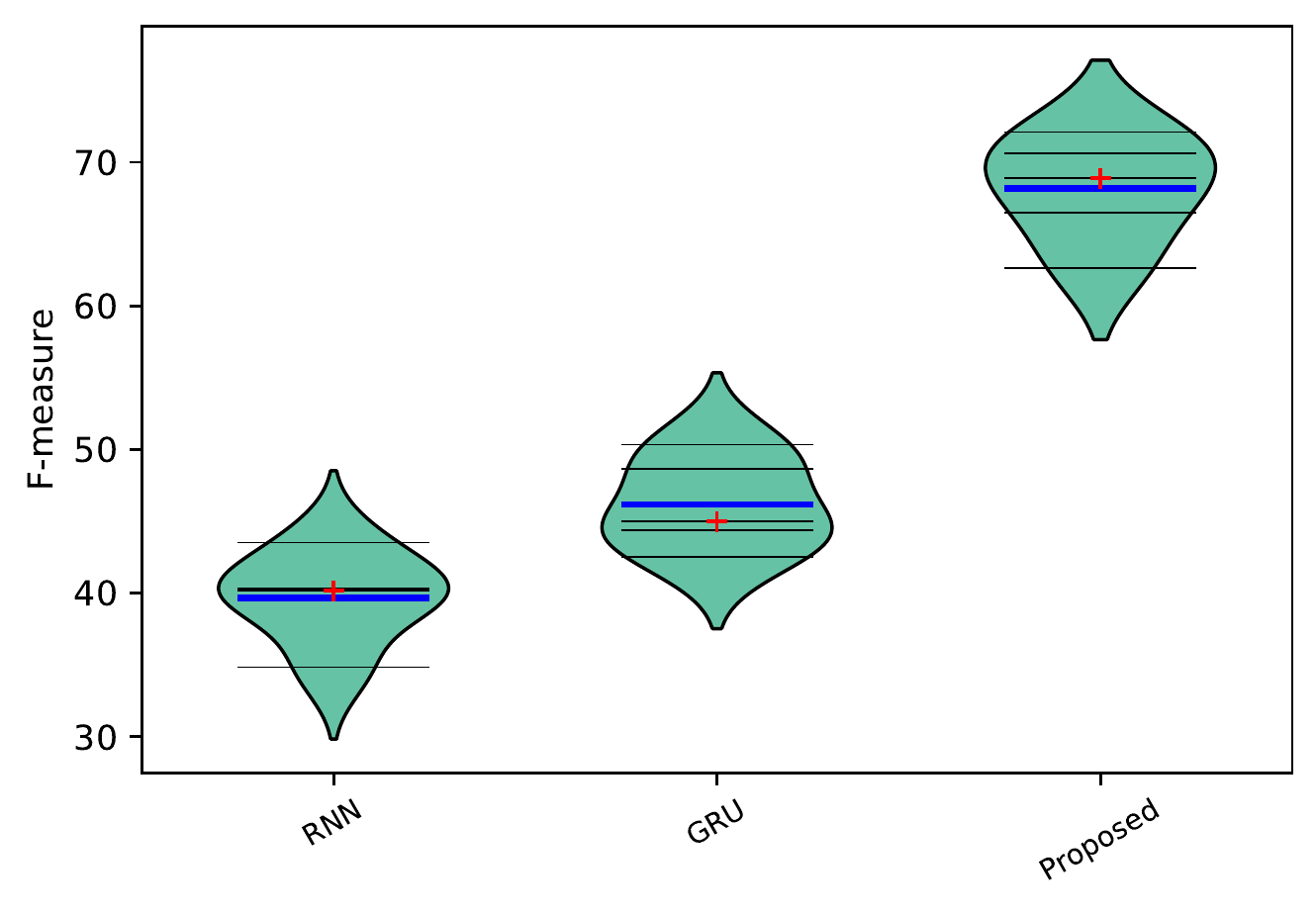}
	\caption{F-measure distribution.}
	\label{fig:f-dist}
\end{figure}

Further, to evaluate the performance of the proposed approach we measure the precision, recall and F-measure scores.
\hyperref[Table:PrecisionScores]{Table} \ref{Table:PrecisionScores} exhibits the precision, recall and F-measure scores. From the \hyperref[Table:PrecisionScores]{Table} \ref{Table:PrecisionScores} and \hyperref[fig:f-dist]{Fig.} \ref{fig:f-dist}, we make the following observations

\begin{itemize}
	\item The proposed approach's average F-measure is \textit{68.36}, while RNN and GRU gain much lower score of \textit{39.73} and \textit{46.20} respectively.
	\item The proposed approach's minimum F-measure is much higher than the maximum F-measure of the baseline approaches.
	\item The results suggest that the proposed approach outperforms the state-of-the-art approaches in precision, recall, and F-measure.
\end{itemize}

\begin{table}[htbp]
	\small
	\caption{Precision, Recall and F-measure comparison with baseline approaches}
	\label{Table:PrecisionScores}
	\begin{center}
		\begin{tabular}{|c|c|cc|c|}
			\toprule
			&  & \multicolumn{2}{c|}{{\underline{     Baselines     }}}  &  \\
			&  & RNN & GRU &  \textbf{Proposed}\\
			\midrule
			\multirow{3}{*}{elasticsearch}  & Precision     & 30.43 & 38.91 & \textbf{63.82} \\
			& Recall     	& 40.75 & 46.86 & \textbf{82.89} \\
			& F-measure     & 34.84 & 42.52 & \textbf{72.11} \\
			\midrule							
			\multirow{3}{*}{spring-boot}  	& Precision     & 37.29 & 40.77 & \textbf{62.39} \\
			& Recall     	& 43.86 & 50.19 & \textbf{62.93} \\
			& F-measure     & 40.31 & 44.99 & \textbf{62.65} \\
			\midrule
			\multirow{3}{*}{RxJava}  		& Precision     & 41.15 & 46.54 & \textbf{66.97} \\
			& Recall     	& 46.20 & 54.81 & \textbf{66.01} \\
			& F-measure     & 43.53 & 50.34 & \textbf{66.48} \\
			\midrule							
			\multirow{3}{*}{java-design} 	& Precision		& 37.68 & 41.23 & \textbf{69.09} \\
			& Recall     	& 41.76 & 48.08 & \textbf{68.71} \\
			& F-measure     & 39.62 & 44.39 & \textbf{68.89} \\
			\midrule							
			\multirow{3}{*}{interviews} 	& Precision     & 38.11 & 47.00 & \textbf{71.02} \\
			& Recall     	& 42.47 & 50.39 & \textbf{70.19} \\
			& F-measure     & 40.17 & 48.63 & \textbf{70.60} \\
			\midrule
			\multirow{3}{*}{\textbf{Average}}& Precision     & 36.93 & 42.89 & \textbf{66.66} \\
			& Recall     	& 43.01 & 50.07 & \textbf{70.15} \\
			& F-measure     & 39.73 & 46.20 & \textbf{68.36} \\
			\bottomrule
		\end{tabular}
	\end{center}
\end{table}

\subsection{RQ2: Comparative analysis for Source Code Suggestion Task}
To further quantify the accuracy of the proposed approach for source code suggestion task, we measure the \textit{Mean Reciprocal Rank (MRR)} scores of each model. The MRR is a rank-based evaluation metric which produces a value between \textit{0-1}, where the value closer to \textit{1} indicates a better source code suggestion model. The MRR can be expressed as

\begin{equation}
MRR(C) = \dfrac{1}{|{C}|}\sum_{i=1}^{|{C}|}\dfrac{1}{y^i}
\end{equation}

where ${C}$ is code sequence and $y^i$ refers to the index of the first relevant prediction. $MRR(C)$ is the average of all sequences $C$ in the test data set.

The results of all models are presented in the \hyperref[Table:MRRScores]{Table} \ref{Table:MRRScores}. The average MRR score of RNN is \textit{0.5156} and the average score of GRU is \textit{0.5749}, while the average score of the proposed approach is \textit{0.7618} which is much higher. From the results, we conclude that the proposed approach significantly outperforms the baseline approaches.

\begin{table}[htbp]
	\small
	\caption{MRR scores with and without proposed approach}
	\label{Table:MRRScores}
	\begin{center}
		\begin{tabular}{|c|cc|c|}
			\toprule	
			&  \multicolumn{2}{c|}{{\underline{     Baselines     }}}  &  \\
			&  RNN & GRU &  \textbf{Proposed}\\
			\midrule
			elasticsearch  	& 0.4851 & 0.5405 & \textbf{0.7344} \\
			\midrule
			spring-boot  	& 0.5161 & 0.5672 & \textbf{0.7363} \\
			\midrule
			RxJava  		& 0.5403 & 0.6085 & \textbf{0.7619} \\
			\midrule
			java-design		& 0.5082 & 0.5625 & \textbf{0.7805} \\
			\midrule
			interviews  	& 0.5284 & 0.5960 & \textbf{0.7960} \\
			\midrule
			\textbf{Average}& 0.5156 & 0.5749 & \textbf{0.7618} \\
			\bottomrule
		\end{tabular}
	\end{center}
\end{table}

To further validate the statistical significance, we employ the ANOVA One-Way statistical test. We conduct the AVOVA test with its default settings ($\alpha$ = 0.05) using Microsoft Excel, and no modifications were made. Comparing (\hyperref[Table:ANOVA]{Table} \ref{Table:ANOVA}) the proposed approach with the best baseline (GRU), we found \textit{ F $>$ F-crit} and \textit{P-value $<$ $\alpha$} is true in all cases (Accuracy, MRR, Precision, Recall and F-measure); therefore, we reject the null hypothesis, suggesting that using different approaches has statistically significant difference in performance.

\begin{table*}[htbp]
	\small
	\centering
	\caption{ANOVA Analysis.}
	\begin{tabular}{lrrrrrr}
		\toprule
		\multicolumn{1}{c}{\textit{Source }} & \multicolumn{1}{c}{\textit{SS}} & \multicolumn{1}{c}{\textit{df}} & \multicolumn{1}{c}{\textit{MS}} & \multicolumn{1}{c}{\textit{F}} & \multicolumn{1}{c}{\textit{P-value}} & \multicolumn{1}{c}{\textit{F-crit}} \\
		\midrule
		
		\textit{Accuracy (K@1)} &       &       &       &       &       &  \\
		Between Groups & 646.416 & 1     & 646.416 & 64.04975 & 4.35463E-05 & 5.317655 \\
		Within Groups & 80.73924 & 8     & 10.092405 &       &       &  \\
		Total & 727.1552 & 9     &       &       &       &  \\
		
		&       &       &       &       &       &  \\
		
		\textit{MRR} &       &       &       &       &       &  \\
		Between Groups & 646.416 & 1     & 646.416 & 64.04975 & 4.35463E-05 & 5.317655 \\
		Within Groups & 80.73924 & 8     & 10.092405 &       &       &  \\
		Total & 727.1552 & 9     &       &       &       &  \\
		
		&       &       &       &       &       &  \\
		
		\textit{Precision} &       &       &       &       &       &  \\
		Between Groups & 1412.295 & 1     & 1412.29456 & 108.0003 & 6.36396E-06 & 5.317655 \\
		Within Groups & 104.6141 & 8     & 13.07676 &       &       &  \\
		Total & 1516.909 & 9     &       &       &       &  \\
		
		&       &       &       &       &       &  \\
		
		\textit{Recall} &       &       &       &       &       &  \\
		Between Groups & 1008.016 & 1     & 1008.016 & 29.81202 & 0.000601504 & 5.317655 \\
		Within Groups & 270.4992 & 8     & 33.812405 &       &       &  \\
		Total & 1278.515 & 9     &       &       &       &  \\
		
		&       &       &       &       &       &  \\
		
		\textit{F-measure} &       &       &       &       &       &  \\
		
		Between Groups & 1206.92 & 1     & 1206.92196 & 99.9581 & 8.5015E-06 & 5.31766 \\
		Within Groups & 96.5942 & 8     & 12.07428 &       &       &  \\
		Total & 1303.52 & 9     &       &       &       &  \\

		\bottomrule
	\end{tabular}%
	\label{Table:ANOVA}%
	\\
	Where, SS = sum of squares, df = degree of freedom, MS = mean square.
\end{table*}%

\subsection{RQ3: Impact of Normalization} 
The evaluation results of the proposed approach for normalized source code and non-normalized source code are presented in \hyperref[Table:Normalization]{Table} \ref{Table:Normalization}. We only remove the comments while building the non-normalized source code. From the results, we observe that the normalization of source code improves the model performance significantly. On average the proposed approach with normalization achieves the accuracy score of 66.15@k=1 where without normalization the accuracy drops to 56.27@k=1. From the results (\hyperref[Table:Normalization]{Table} \ref{Table:Normalization}), we conclude that the normalization process significantly affects the model performance.

\begin{table}[htbp]
	\small
	\caption{Impact of Normalization}
	\label{Table:Normalization}
	\begin{center}
		\begin{tabular}{|c|c|c|c|c|c|}
			\toprule
			&  \textbf{Accuracy} & \textbf{Precision} & \textbf{Recall} & \textbf{F-measure} & \textbf{MRR}\\
			\midrule
			Normalized  	& 66.15 & 66.66 & 70.15 & 68.36 & 0.7618 \\
			\midrule
			Non-Normalized  & 56.27 & 57.25 & 62.14	& 54.66	& 0.6524 \\
			\bottomrule
		\end{tabular}
	\end{center}
\end{table}

\subsection{Additional Findings}
\begin{figure}[h]
	\centering
	\includegraphics[width=0.8\linewidth]{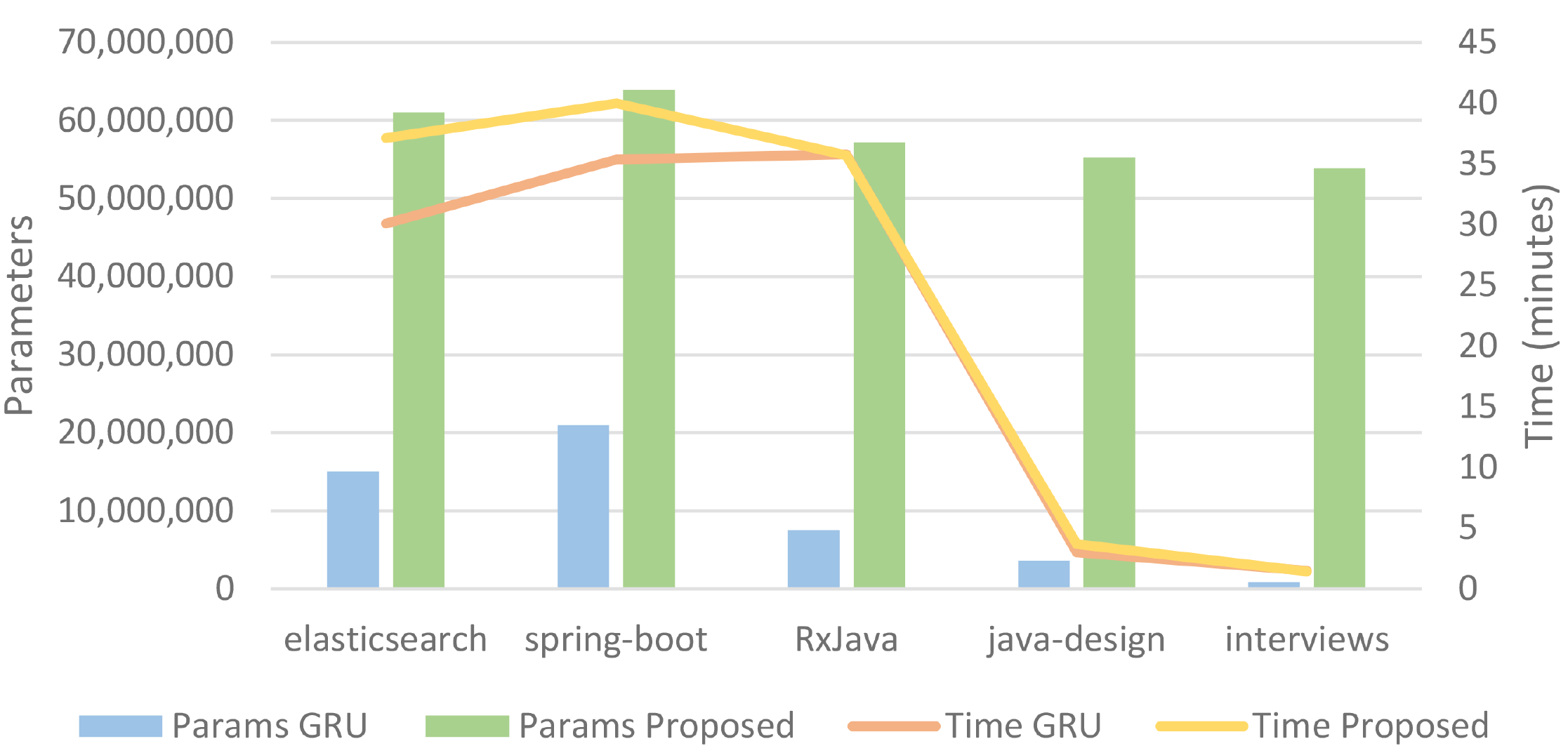}
	\caption{Model parameters and train time (epoch).}
	\label{fig:prams}
\end{figure}

From our experiments, we find several interesting facts. First, we notice that the transfer learning-based models take advantage of a large set of pre-trained parameters resulting in a significant performance boost. \hyperref[fig:prams]{Fig.} \ref{fig:prams} shows the parameters in each model and time per epoch. we can observe that the proposed approach's average parameters are 58M (million) which are much higher as compared to other baseline's average parameters which are 9M. The parameters space of the proposed approach is much larger than the other baseline with minimum to none overhead on time. The proposed approach significantly boosts the model's performance by leveraging pre-trained knowledge without needing to learn the parameters from scratch. One important thing to mention here is that the model training is offline thus has no impact on the source code suggestion task. The proposed approach can suggest the next source code token in less than 20 milliseconds. Moreover, we experimented with another variant of the recurrent neural network named LSTM. We found that the performance of LSTM is worse as compared to RNN and GRU, thus we choose not to use it for transfer learning purpose. Furthermore, to evaluate the proposed approach qualitatively consider the example input code \inlinecode{public static void display(int[][] matrix) \{ System.out.}, where the next possible source code token could be \inlinecode{println}. The proposed approach correctly captures the source code context and predicts the most probable next source code suggestions \inlinecode{[println,print,writeShort]}, effectively ranking \inlinecode{println} on its first index.

The proposed approach attains the finest performance due to several different reasons. First, the proposed approach takes leverage from pre-trained models by transferring the learned features from them. Second, the \textit{attention} learner fine-tunes the model by paying attention to only task-specific features and does not increase the computational complexity which resulted in better performance. Consequently, transfer learning-based \textit{attention} model has better generalization capability without training the model from scratch.

The broader impact of our work is to show that transfer learning could be beneficial in the domain of source code modeling. This work is the first step in this direction and results encourage future research on it. The work can be improved in several different ways. First, the performance of the proposed approach can be improved by hyper-parameter optimization \cite{matuszyk2016comparative}. Second, the proposed approach can be improved by using complex architectures such as transformers \cite{devlin2018bert} and stacked neural networks \cite{vincent2010stacked}. Another possible path for improvement is to train the model on an even larger data set. In the future, we consider exploiting these possibilities.

\section{Threats to Validity}

A risk to construct validity is the selection of assessment metrics. To alleviate this threat, we use several different evaluation metrics. We use the Top-k accuracy metric as done by former studies 
\cite{hindle2012naturalness,white2015toward,nguyen2018deep}. We use the precision, recall, and F-measure \cite{alon2019code2vec} metrics for the evaluation of the proposed approach. These metrics are most generally used for the model evaluation purpose. Moreover, we evaluate the proposed approach with MRR \cite{nguyen2018deep,santos2018syntax} metric which is a ranked based metric. Further, to show the statistical significance of the proposed approach we adopt the ANOVA statistical testing.  

A risk to internal validity is the employment of the baseline methods. We re-implement the baseline approaches by following the process described in the original manuscripts. To alleviate this risk, we twofold the implementations and results. Conversely, there could be some unobserved inaccuracies. Another risk is the choice of hyper-parameters for deep learning methods. The change in training, validation or testing set or the variation in hyper-parameters may impact the performance of the anticipated method.

A threat to external validity is related to the generality of results. The data set used in this study is collected from \textit{GitHub}, a well-known source code repositories provider. It is not necessary that the projects used in this study represent other languages or Java language source code entirely.

\section{Conclusion}
In this work, we proposed a deep learning-based source code language model by using the concept of transfer learning. First, we exploit the concept of transfer learning for neural language-based source code models. Next, we presented RNN and GRU based pre-trained models for the purpose of transfer learning in the domain of source code. Both models are trained on over 13 million code tokens and do not need retraining and can directly be used for the purpose of transfer learning. We evaluated the proposed approach with the downstream task of source code suggestion. We evaluated the proposed approach extensively and compared it with the state-of-the-art models. The extensive evaluation of this work suggests that the proposed approach significantly improves the model's performance by exploiting the concept of transfer learning.

\section*{Acknowledgments}
This work was supported by the National Key R\&D (grant no. 2018YFB1003902), Natural Science Foundation of Jiangsu Province (No. BK20170809), National Natural Science Foundation of China (No. 61972197) and Qing Lan Project.

\bibliography{TransferLearning}

\end{document}